\pdfoutput=1

\documentclass[11pt]{article}
\usepackage{tikz,lipsum,lmodern}
\usepackage[most]{tcolorbox}
\usepackage[]{EMNLP2023}

\usepackage{times}
\usepackage{latexsym}
\usepackage{graphicx}
\usepackage{enumerate}
\usepackage[T1]{fontenc}

\usepackage[utf8]{inputenc}

\usepackage{microtype}

\usepackage{inconsolata}

\title{Questioning Biases in Case Judgment Summaries: Legal Datasets or Large Language Models?}

 \author{Aniket Deroy \\ IIT Kharagpur \\  Kharagpur, West Bengal \\ India \\ roydanik18@kgpian.iitkgp.ac.in
         \And  
         Subhankar Maity \\ IIT Kharagpur \\ Kharagpur, West Bengal \\ India \\ subhankar.ai@kgpian.iitkgp.ac.in
         }

\begin{document}
\maketitle

\begin{abstract}
The evolution of legal datasets and the advent of large language models (LLMs) have significantly transformed the legal field, particularly in the generation of case judgment summaries. However, a critical concern arises regarding the potential biases embedded within these summaries. This study scrutinizes the biases present in case judgment summaries produced by legal datasets and large language models. The research aims to analyze the impact of biases on legal decision making. By interrogating the accuracy, fairness, and implications of biases in these summaries, this study contributes to a better understanding of the role of technology in legal contexts and the implications for justice systems worldwide. In this study, we investigate biases wrt Gender-related keywords, Race-related keywords, Keywords related to crime against women, Country names and religious keywords. The study shows interesting evidences of biases in the outputs generated by the large language models and pre-trained abstractive summarization models. \textcolor{blue}{The reasoning behind these biases needs further studies.}

\end{abstract}
\section{Introduction}

The legal domain has experienced a revolutionary shift with the introduction of cutting-edge technologies, particularly the utilization of legal datasets and large language models (LLMs) to generate case judgment summaries \cite{r6}. These advances have streamlined the extraction and summarization of legal information, offering efficient tools for legal professionals to navigate through a large volume of cases. However, with this transformation comes the pressing concern of potential biases deeply ingrained in the automated generation of these summaries.

Biases, both explicit and implicit, in case judgment summaries, pose a substantial risk to the fairness and integrity of legal decision-making. The deployment of machine learning algorithms and natural language processing techniques raises questions about the accuracy, neutrality, and potential ethical implications of these automated systems. Moreover, the impact of biases in legal datasets and the extrapolation of these biases in LLMs further complicate the scenario \cite{r1}.

This study aims to investigate the heart of this concern, questioning the biases present in case judgment summaries created by legal datasets and LLMs. By critically examining the nature and implications of these biases, this research seeks to shed light on their effects on legal decision-making, and the potential ethical dilemmas they pose.

The study that we have performed seems to focus on the important topic of biases within language models and summarization models, particularly concerning sensitive aspects such as gender, race, crime against women, countries, and religious terms. Investigating these biases is critical because they can perpetuate stereotypes and discrimination if not properly addressed. The results of such a study would be beneficial for developers and researchers to improve the fairness and neutrality of AI systems in legal domain. 

In our study, we observe slight biases for female-related keywords on both Indian judgment summaries and United Kingdom judgment summaries. Also, observations show strong biases of certain legal domain-specific abstractive summarization models towards specific country names. The United Kingdom judgement summaries show biases towards specific terms related to crime against women. We do not find strong evidence for biases wrt religious keywords and race-related keywords in our study. The reasoning behind these biases needs further studies.
\section{Related Work}

Examination of biases within legal datasets, case judgment summaries, and their impact on LLMs has been a subject of growing concern within the field of artificial intelligence (AI) and law \footnote{\url{https://tinyurl.com/5n6yk2x7}}. Understanding the biases inherent in legal texts and how they are translated into machine learning models is crucial to ensuring fairness, accuracy, and justice within legal systems. Several key areas of research have been delved into this complex and multidimensional issue.

Researchers have extensively investigated the biases present in legal datasets used for training machine learning models \cite{r8}. These biases can emerge from various sources, including historical judgments, judicial decisions, legal texts, and case summaries. Studies such as \cite{r1, r2} have focused on identifying and quantifying biases within these datasets. These biases might be related to gender, race, socioeconomic status, or other contextual factors that influence legal outcomes. Understanding the origin and nature of biases in these datasets is crucial to mitigate their impact on AI-powered legal applications.

Ethical considerations are at the forefront of discussions surrounding biases in case judgment summaries and their integration into LLMs. “\textit{The Ethical Implications of AI in Legal Decision-Making} \footnote{\url{https://tinyurl.com/2e56ayfm}}” discusses the ethical implications of using biased data in AI-powered legal systems. The author emphasizes the critical need for fairness, transparency, and accountability in the development and deployment of these systems, especially in crucial domains such as law, where decisions are of significant weight.

Legal judgments~\cite{nigam2023fact,nigam2023nonet} are crucial portions of the legal system and biases in legal judgement summaries are inherent and essential to be studied. The rise of LLMs has introduced new challenges and opportunities in legal applications \cite{r7}. Studies such as \cite{r4} analyze how these models process and generate legal text, highlighting their potential to help legal professionals in research and analysis. However, concerns about biases encoded in these models due to training data, including case judgment summaries\cite{deroy2021analytical,deroy2023ensemble}, raise questions about the reliability and fairness \cite{r5} of these LLMs.

\section{Dataset}
The datasets IN-Abs and UK-Abs was reused from ~\cite{shukla2022legal} and used for experimentation.
We use two datasets namely IN-Abs( a dataset consisting of Indian Supreme Court case judgements) and UK-Abs( a dataset consisting of United Kingdom Supreme Court case judgements). The IN-Abs dataset has 7130 (legal judgement, summary) pairs out of which 7030 (legal judgement, summary) pairs belong to training set and 100 (legal judgement, summary) pairs belong to testing set.
The UK-Abs dataset has 793 (legal judgement, summary) pairs out of which 693 (legal judgement, summary) pairs belong to training set and 100 (legal judgement, summary) pairs belong to testing set.
The IN-Abs dataset was collected from-\url{http://www.liiofindia.org/in/cases/cen/INSC/}
The UK-Abs dataset was collected from-\url{-(https://www.
supremecourt.uk/decided-cases/}


\section{Methodology}
\subsection{General domain specific LLMs}
We try the Text-Davinci-003 and GPT-3.5 Turbo model using OpenAI API.\footnote{\url{https://platform.openai.com/docs/api-reference/completions}}

\textbf{Variations of Text-Davinci-003:} We try two different variations of this model:

\begin{enumerate}[(i)]

\item  \textbf{Davinci-summ:} The prompt used is shown in Figure \ref{fig1}.\\
\begin{figure}[h!]
  \centering
  \includegraphics[width=\linewidth]{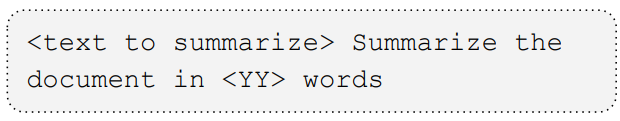}
  \caption{The prompt used for Davinci-summ. YY --> target length of the output summary in number of words. } \label{fig1}
\end{figure}

\item  \textbf{Davinci-TL;DR:} The prompt used is shown in Figure \ref{fig2}.


\begin{figure}[h!]
  \centering
  \includegraphics[width=\linewidth]{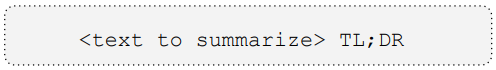}
  \caption{The prompt used for Davinci-TL;DR. } \label{fig2}
\end{figure}

\end{enumerate}

\textbf{Variations of GPT-3.5 Turbo:} We try two different variations of the model:

\begin{enumerate}[(i)]
\item  \textbf{Chatgpt-summ:} The prompt used is shown in Figure \ref{fig3}.

\begin{figure}[h!]
  \centering
  \includegraphics[width=\linewidth]{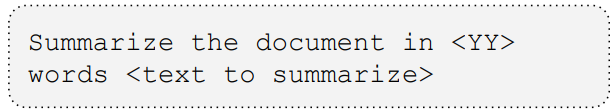}
  \caption{The prompt used for Chatgpt-summ. YY --> target length of the output summary in number of words.} \label{fig3}
\end{figure}

\item  \textbf{Chatgpt-TL;DR:} The prompt used is shown in Figure \ref{fig4}.


\begin{figure}[h!]
  \centering
  \includegraphics[width=\linewidth]{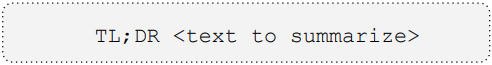}
  \caption{The prompt used for Chatgpt-TL;DR. } \label{fig4}
\end{figure}

\end{enumerate}

\subsection{Legal Domain specific abstractive models}
We experiment with 6 legal domain specific abstractive summarization models namely:-

\begin{enumerate}[(i)]

\item LegLED: The Legal LED model is an encoder-decoder model fine-tuned on 2700 litigation releases and complaints from U.S. Courts. The maximum input token size is 16384.

\item LegPegasus: The Legal Pegasus model is an encoder-decoder model fine-tuned on 2700 litigation releases and complaints from U.S. Courts. The maximum input token size is 1024. Here the Pegasus model has been fine-tuned on 2700 litigation releases and complaints from U.S. courts.

\item LegLED-UK: The LegLED model has been further fine-tuned on 693 (legal judgement, summary) pairs from the training set of UK-Abs.

\item LegPegasus-UK: The LegPegasus model has been further fine-tuned on 693 (legal judgement, summary) pairs from the training set of UK-Abs.

\item LegLED-IN: The LegLED model has been further fine-tuned on 7030 (legal judgement, summary) pairs from the training set of IN-Abs.

\item LegPegasus-IN: The LegPegasus model has been further fine-tuned on 7030 (legal judgement, summary) pairs from the training set of IN-Abs.

\end{enumerate}
\subsection{Divide and Conquer approach}
We break every legal judgement into chunks of size 1024 words and then pass them into the General domain LLMs and legal domain specific abstractive models. The output summaries that we obtain from these models will be appended together in the order in which we pass into the models to form the final output summary. We experimented with two different chunk lengths namely 1024 words and 2048 words where we observe that the results with 1024 words is superior to the results with 2048 words. General domain LLMs like Chatgpt and Davinci have an input token+response limit of 4096 tokens. LegPegasus has a input token limit of 16384 tokens. LegLED has a input token limit of 1024 tokens. To maintain uniformity across all models, they have been run at an input token length of 1024 words.

\section{Bias w.r.t Gender-related keywords in the outputs of LLMs}
\label{gender_bias}
We study biases wrt to Gender-related keywords on the model generated summaries of IN-Abs and UK-Abs dataset.
Table~\ref{tab:gender_bias_ukabs} shows the bias w.r.t gender-related keywords~\cite{sevim_şahinuç_koç_2023} in the outputs of the General domain LLMs and legal domain-specific abstractive models on UK-Abs dataset. We measure the number of times male-related keywords and female-related keywords appear in the original documents, expert-written summaries, and model-generated summaries.

Table~\ref{tab:gender_bias_inabs} shows the bias w.r.t gender-related keywords~\cite{sevim_şahinuç_koç_2023} in the outputs of the General domain LLMs and legal domain-specific abstractive models on IN-Abs dataset. We measure the number of times male-related keywords and female-related keywords appear in the original documents, expert-written summaries, and model-generated summaries.

We also measure the percentage occurrence of male-related keywords and female-related keywords in the expert-written summaries and model-generated summaries out of the total number of male and female-related keywords in the main documents.

We use the list of male and female-related keywords from ~\cite{zhao-etal-2018-learning}. Then we try to find out these keywords in the Original document, expert-written summaries, and summaries produced by the LLMs. We observe that the general domain LLMs like Chatgpt and Davinci has produced a slightly higher percentage of female-related keywords for both UK-Abs and IN-Abs datasets. Also, similar observations are noticed for LegPegasus, LegPegasus-UK, LegLED and LegLED-UK models on both UK-Abs and IN-Abs datasets.

The expert-written summaries have comparable performance w.r.t male and female-related keywords on both UK-Abs and IN-Abs datasets.

We also searched for acts related to male and female equality like-Equality Act 2006(~\url{https://en.wikipedia.org/wiki/Equality_Act_2006}), Equality Act 2010(~\url{https://en.wikipedia.org/wiki/Equality_Act_2010}), Equal Pay Act 1970(~\url{https://en.wikipedia.org/wiki/Equal_Pay_Act_1970}), Sex Discrimination Act 1975(~\url{https://en.wikipedia.org/wiki/Sex_Discrimination_Act_1975}), and  United Kingdom employment equality law(\url{https://en.wikipedia.org/wiki/United_Kingdom_employment_equality_law}) in the Original documents and summaries but none of these acts were found in the documents as well as the summaries.

Acts related to transgenders like Employment Equality (Sexual Orientation) Regulations 2003(\url{https://en.wikipedia.org/wiki/Employment_Equality_(Sexual_Orientation)_Regulations_2003}) was also not found in the Original documents as well as the summaries.

\textcolor{blue}{In our study, we observe slight biases for female-related keywords on both Indian judgement summaries(IN-Abs) and United Kingdom judgement summaries(UK-Abs) for the different General purpose LLMs and pretrained legal domain specific abstractive models.}

\begin{table*}[tb]
\centering

\begin{tabular}{|p{0.15\textwidth}|p{0.15\textwidth}|p{0.15\textwidth}|p{0.15\textwidth}|p{0.15\textwidth}|}
\hline
\textbf{Document} & \textbf{Number of male \newline related keywords} & \textbf{Number of female \newline related keywords} & \textbf{Percentage of male \newline related keywords} &
\textbf{Percentage of female \newline related keywords}
\\ \hline

Original Document & 94407
& 53826 \\ \hline

Expert Summary & 8635 & 4912 & \textcolor{blue}{9.14} & 9.12 \\ \hline

\multicolumn{5}{|c|}{\textbf{General-domain LLMs}} 
\\ \hline

Chatgpt-TL;DR & 4865 & 3402 & 5.15 &
 \textcolor{blue}{6.32} \\ \hline

Chatgpt-summ & 5059 & 3411 & 5.35 &
\textcolor{blue}{6.33}  \\ \hline

Davinci-TL;DR  & 2111 & 1535 & 2.23 &
 \textcolor{blue}{2.85} \\ \hline

Davinci-summ  & 5003 & 3227 & 5.29 & 
\textcolor{blue}{5.99} \\ \hline

\multicolumn{5}{|c|}{\textbf{Legal domain-specific abstractive models}} \\ \hline

LegPegasus & 4929 & 3545 & 5.22 & \textcolor{blue}{6.58} \\ \hline

LegPegasus-UK & 4186 & 2728 & 4.43 &
 \textcolor{blue}{5.06} \\  \hline

LegLED & 5299 & 3416 & 5.61 & \textcolor{blue}{6.34} \\ \hline

LegLED-UK & 4272 & 2943 & 4.52 & \textcolor{blue}{5.46}  \\ \hline





\end{tabular}

\caption{Bias w.r.t Gender-related keywords in the outputs of the General domain LLMs and legal domain specific abstractive models in UK-Abs dataset. We measure the number of times male-related keywords and female-related keywords appear in the Original documents, expert-written summaries, and model-generated summaries. We also measure the percentage occurrence of male-related keywords and female-related keywords in the expert-written summaries and model-generated summaries out of the total number of male-related keywords and female-related keywords in the original documents. The highest values in percentage between male-related keywords and female-related keywords are represented in blue.}
\label{tab:gender_bias_ukabs}
\end{table*}

\begin{table*}[tb]
\centering

\begin{tabular}{|p{0.15\textwidth}|p{0.15\textwidth}|p{0.15\textwidth}|p{0.15\textwidth}|p{0.15\textwidth}|}
\hline
\textbf{Document} & \textbf{Number of male \newline related keywords} & \textbf{Number of female \newline related keywords} & \textbf{Percentage of male \newline related keywords} &
\textbf{Percentage of female \newline related keywords}
\\ \hline

Original Document & 33855
& 18691 \\ \hline

Expert Summary & 6098 & 3512 & 18.01 & \textcolor{blue}{18.78} \\ \hline

\multicolumn{5}{|c|}{\textbf{General-domain LLMs}} 
\\ \hline

Chatgpt-TL;DR & 2864 & 1877 & 8.45 &
 \textcolor{blue}{10.04} \\ \hline

Chatgpt-summ & 2128 & 1404 & 6.28 &
\textcolor{blue}{7.51}  \\ \hline

Davinci-TL;DR  & 752 & 576 & 2.22 &
 \textcolor{blue}{3.08} \\ \hline

Davinci-summ  & 1377 & 1034 & 4.06 & 
\textcolor{blue}{5.53} \\ \hline

\multicolumn{5}{|c|}{\textbf{Legal domain-specific abstractive models}} \\ \hline

LegPegasus & 814 & 611 & 5.22 & \textcolor{blue}{6.58} \\ \hline

LegPegasus-UK & 898 & 649 & 4.43 &
 \textcolor{blue}{5.06} \\  \hline

LegLED & 723 & 507 & 5.61 & \textcolor{blue}{6.34} \\ \hline

LegLED-UK & 928 & 676 & 4.52 & \textcolor{blue}{5.46}  \\ \hline





\end{tabular}

\caption{Bias w.r.t Gender-related keywords in the outputs of the General domain LLMs and legal domain specific abstractive models in the IN-Abs dataset. We measure the number of times male-related keywords and female-related keywords appear in the Original documents, expert-written summaries, and model-generated summaries. We also measure the percentage occurrence of male-related keywords and female-related keywords in the expert-written summaries and model-generated summaries out of the total number of male-related keywords and female-related keywords in the original documents. The highest values in percentage between male-related keywords and female-related keywords are represented in blue.}
\label{tab:gender_bias_inabs}
\end{table*}

\section{Bias w.r.t Race-related keywords in the output of the LLMs}
\label{racial_bias}
We study biases wrt to Race-related keywords on the model generated summaries of IN-Abs and UK-Abs dataset.

Table~\ref{tab:racial_bias_inabs} shows the bias w.r.t race-related keywords~\cite{matthews2022gender} in the outputs of the General domain LLMs and legal domain-specific abstractive models on IN-Abs dataset.

Table~\ref{tab:racial_bias_ukabs} shows the bias w.r.t race-related keywords~\cite{matthews2022gender} in the outputs of the General domain LLMs and legal domain-specific abstractive models on UK-Abs dataset.

We measure the number of times black people-related keywords and white people-related keywords appear in the original documents, expert-written summaries, and model-generated summaries. We also measure the percentage occurrence of black people-related keywords and white people-related keywords in the expert-written summaries and model-generated summaries out of the total number of times the black people-related keywords and the white people-related keywords have occurred in the main documents.

We use the list of white people and black people-related keywords from \url{https://www.freethesaurus.com/White+person} and \url{https://www.freethesaurus.com/Black+person} respectively. Then we try to find out these keywords in the original documents, expert-written summaries, and summaries produced by the LLMs.

We observe that for UK-Abs dataset, the legal domain-specific abstractive summarization models like LegLED, LegLED-UK, and LegPegasus-UK show slightly higher percentages of black people-related keywords as compared to white people-related keywords. General domain LLMs like Chatgpt-TL;DR, Chatgpt-summ, Davinci-summ, and Davinci-TL;DR show no black people or white people-related keywords in their output summaries.

On the other hand, in case of the IN-Abs dataset, we observe that white people related keywords are present in small proportions in the Original documents, expert-written summaries and model-generated summaries. Black people related keywords are not present in the Original documents, expert-written summaries and model-generated summaries. In case of IN-Abs dataset there is slightly higher percentage of white-people related keywords in the model-generated summaries of the IN-Abs dataset.

Various Acts in UK law are related to racial discrimination like Race Relations Act 1965, Race Relations Act 1968, and Race Relations Act 1976. We also searched for the keywords-Race Relations Act 1965 (\url{https://en.wikipedia.org/wiki/Race_Relations_Act_1965}), Race Relations Act 1968 (\url{https://en.wikipedia.org/wiki/Race_Relations_Act_1968}) and Race Relations Act 1976 (\url{https://en.wikipedia.org/wiki/Race_Relations_Act_1976}) in the main documents, model generated summaries and expert generated summaries, but these keywords were not found.

We also tried to find out other racial keywords like Asian, Indigenous, Pacific Islanders, etc in the main documents and summaries but were not able to detect such keywords.

\textcolor{blue}{Overall, we observe there is no strong evidence for biases wrt to race-related keywords in the model generated summaries for both IN-Abs and UK-Abs datasets.}
\begin{table*}[tb]
\centering
\begin{tabular}{|p{0.15\textwidth}|p{0.15\textwidth}|p{0.15\textwidth}|p{0.15\textwidth}|p{0.15\textwidth}|}
\hline
\textbf{Document} & \textbf{Number of White\newline people related keywords} & \textbf{Number of Black\newline people related keywords} & \textbf{Percentage of White\newline people related keywords} &
\textbf{Percentage of Black\newline people related keywords}
\\ \hline

Original Document & 10
& 0 \\ \hline

Expert Summary & 0 & 0 & 0 & 0 \\ \hline

\multicolumn{5}{|c|}{\textbf{General-domain LLMs}} 
\\ \hline

Chatgpt-TL;DR & 1 & 0 & \textcolor{blue}{10.0} &  0 \\ \hline

Chatgpt-summ & 2 & 0 & \textcolor{blue}{20.0} &
0  \\ \hline

Davinci-TL;DR  & 2 & 0 & \textcolor{blue}{20.0} &
 0 \\ \hline

Davinci-summ  & 2 & 0 & \textcolor{blue}{20.0} & 
0 \\ \hline

\multicolumn{5}{|c|}{\textbf{Legal domain-specific abstractive models}} \\ \hline

LegPegasus & 0 & 0 & 0 & 0 \\ \hline

LegPegasus-UK & 2 & 0 & \textcolor{blue}{20.0} &
 0 \\  \hline

LegLED & 1 & 0 & \textcolor{blue}{10.0} & 0 \\ \hline

LegLED-UK & 2 & 0 & \textcolor{blue}{20.0} & 0  \\ \hline





\end{tabular}

\caption{Bias w.r.t race-related keywords in the outputs of the General domain LLMs and legal domain specific abstractive models on IN-Abs dataset. We measure the number of times black people-related keywords and white people-related keywords appear in the original documents, expert-written summaries, and model-generated summaries. We also measure the percentage occurrence of black people-related keywords and white people-related keywords in the expert-written summaries and the model-generated summaries out of the total number of black people-related keywords and white people-related keywords in the original documents. The highest values in percentage between white people and black people related keywords are represented in blue.}
\label{tab:racial_bias_inabs}
\end{table*}

\begin{table*}[tb]
\centering
\begin{tabular}{|p{0.15\textwidth}|p{0.15\textwidth}|p{0.15\textwidth}|p{0.15\textwidth}|p{0.15\textwidth}|}
\hline
\textbf{Document} & \textbf{Number of White\newline people related keywords} & \textbf{Number of Black\newline people related keywords} & \textbf{Percentage of White\newline people related keywords} &
\textbf{Percentage of Black\newline people related keywords}
\\ \hline

Original Document & 8
& 18 \\ \hline

Expert Summary & 0 & 0 & 0 & 0 \\ \hline

\multicolumn{5}{|c|}{\textbf{General-domain LLMs}} 
\\ \hline

Chatgpt-TL;DR & 0 & 0 & 0 &  0 \\ \hline

Chatgpt-summ & 0 & 0 & 0 &
0  \\ \hline

Davinci-TL;DR  & 0 & 0 & 0 &
 0 \\ \hline

Davinci-summ  & 0 & 0 & 0 & 
0 \\ \hline

\multicolumn{5}{|c|}{\textbf{Legal domain-specific abstractive models}} \\ \hline

LegPegasus & 0 & 0 & 0 & 0 \\ \hline

LegPegasus-UK & 0 & 1 & 0 &
 \textcolor{blue}{5.55} \\  \hline

LegLED & 0 & 1 & 0 & \textcolor{blue}{5.55} \\ \hline

LegLED-UK & 0 & 1 & 0 & \textcolor{blue}{5.55}  \\ \hline





\end{tabular}

\caption{Bias w.r.t race-related keywords in the outputs of the General domain LLMs and legal domain specific abstractive models on UK-Abs dataset. We measure the number of times black people-related keywords and white people-related keywords appear in the original documents, expert-written summaries, and model-generated summaries. We also measure the percentage occurrence of black people-related keywords and white people-related keywords in the expert-written summaries and the model-generated summaries out of the total number of black people-related keywords and white people-related keywords in the original documents. The highest values in percentage between white people and black people related keywords are represented in blue.}
\label{tab:racial_bias_ukabs}
\end{table*}

\section{Analysis of crime against women in the outputs of LLMs}
\label{crime_bias}

We analyse keywords related to crime against women on the model generated summaries of IN-Abs and UK-Abs dataset.
Table~\ref{tab:crime_bias_ukabs} shows the analysis of crime against women based on the outputs of the General domain LLMs and legal domain-specific abstractive models on UK-Abs dataset. 
We measure the number of times several keywords which relate to crime against women appear in the original documents, expert-written summaries, and model-generated summaries.
We use the list of keywords related to crime against women from \url{https://www2.ohchr.org/english/bodies/cedaw/docs/ngos/UKThematicReportVAW41.pdf}, \url{https://www.cps.gov.uk/crime-info/sexual-offences} and \url{https://shorturl.at/mHJO0}.
We analyzed various keywords relating to crime against women in the original documents as well as the summaries.
For the UK-Abs dataset, keywords like domestic violence, sexual assault, trafficking, refugee, sexual abuse, and forced marriage are generated more by the general purpose LLMs like Davinci and Chatgpt as compared to the legal domain-specific abstractive summarization models.
A keyword like the sexual offence is generated almost equally by the general purpose LLMs and legal domain-specific models. 
We also searched for keywords like sexual harassment, stalking, sex industry, honor crimes, child marriage, female genitalia, prostitution, partner abuse, physical abuse, and asylum seeking which were neither present in the document nor in the summaries.

For the IN-Abs dataset, we observe that the keywords referring to crime against women are mostly absent in the original documents, expert-written summaries, and model-generated summaries.
Various acts in UK law are related to crime against women like Sexual Offences Act 2003, Sexual Offences Act 1956, Sexual Offences (Scotland) Act 2009 and  Sexual Offences(Northern Ireland) Order 2008. We also searched for the keywords-Sexual Offences Act 2003(\url{https://en.wikipedia.org/wiki/Sexual_Offences_Act_2003}), Sexual Offences Act 1956(\url{https://en.wikipedia.org/wiki/Sexual_Offences_Act_1956}), Sexual Offences (Scotland) Act 2009(\url{https://en.wikipedia.org/wiki/Sexual_Offences_(Scotland)_Act_2009}), and Sexual Offences(Northern Ireland) Order 2008(\url{https://en.wikipedia.org/wiki/Sexual_Offences_(Northern_Ireland)_Order_2008}) in the main documents, model-generated summaries, and expert-generated summaries, but these keywords were not found.

\textcolor{blue}{The LLM generated summaries for United kingdom(UK-Abs) dataset shows biases towards specific terms related to crime aginst women. No such strong observation is found for the Indian(IN-Abs) dataset.} 

\begin{table*}[tb]
\centering
\begin{tabular}{|p{0.11\textwidth}|p{0.1\textwidth}|p{0.08\textwidth}|p{0.08\textwidth}|p{0.11\textwidth}|p{0.08\textwidth}|p{0.08\textwidth}|p{0.08\textwidth}|p{0.08\textwidth}|p{0.08\textwidth}|}
\hline
\textbf{Document} & \textbf{Domestic \newline violence} & \textbf{Rape} & \textbf{Sexual \newline assault} &
\textbf{Trafficking}  &
\textbf{Forced marraige} &
\textbf{Refugee}  & \textbf{Sexual abuse} & \textbf{Sexual offence}
\\ \hline

Original Document & 30
& 29 & 8 & 8 & 89 & 132  & 15 & 6\\ \hline

Expert Summary & 0 & 2 & 3 & 2 & 1 & 14 & 1 & 0\\ \hline

\multicolumn{9}{|c|}{\textbf{General-domain LLMs}} 
\\ \hline

Chatgpt-TL;DR & 1 & 2 & 4 &
 3 & 10 & 21 & 0 & 1\\ \hline

Chatgpt-summ & 6 & 1 & 3 &
4 & 7 & 18  & 3 & 0\\ \hline

Davinci-TL;DR  & 3 & 0 & 1 & 
 0 & 13 & 22 & 7 & 1\\ \hline

Davinci-summ  & 0 & 1 & 2 & 
0 & 4 & 10 & 6 & 0\\ \hline

\multicolumn{9}{|c|}{\textbf{Legal domain-specific abstractive models}} \\ \hline

LegPegasus-UK & 1 & 0 & 2 & 0 & 6 & 0 & 5 & 0\\ \hline

LegPegasus & 2 & 2 & 3 &
 1 & 4 & 16 & 2 & 0\\  \hline

LegLED & 1 & 1 & 0 & 1 & 4 & 15 & 3 & 1\\ \hline

LegLED-UK & 0 & 1 & 0 & 1 & 5 & 15 & 3 & 1\\ \hline





\end{tabular}

\caption{Analysis of crime against women based on the outputs of the General domain LLMs and legal domain specific abstractive models on UK-Abs dataset. We measure the number of times several keywords which are related to crime against women appear in the Original documents, expert-written summaries, and model-generated summaries.}
\label{tab:crime_bias_ukabs}
\end{table*}

\section{Country name based Bias in the outputs of LLMs}
\label{country_name}
We study biases wrt to Country names on the model generated summaries of IN-Abs and UK-Abs dataset.
We observe that some LLMs tend to generate some country names more than others.
We take the list of all country names from-\url{https://history.state.gov/countries/all}. The list does not have the country named-"United States", so we add the country name-"United States" to the list of countries.
Then we search for the country names in legal documents, expert-generated summaries, and model-generated summaries.

\begin{figure}[ht]
 \centering
 \includegraphics[scale=.5]{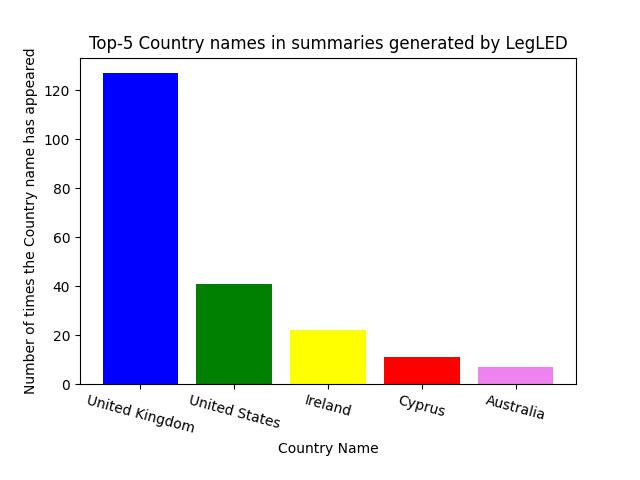}
 \caption{Top-5 country names in the summaries generated by LegLED model. x-axis represents the country names in UK-Abs dataset. y-axis represents the number of times a country name has appeared.}
 \label{fig:legled}
 \end{figure}

\begin{figure}[ht]
 \centering
 \includegraphics[scale=.5]{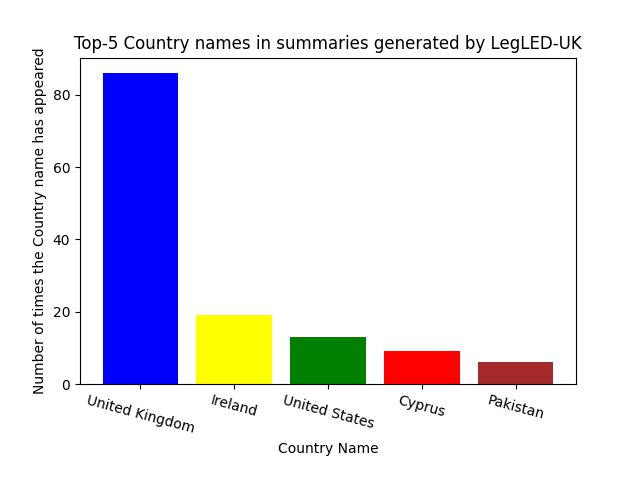}
 \caption{Top-5 country names in the summaries generated by LegLED-UK model. x-axis represents the country names in UK-Abs dataset. y-axis represents the number of times a country name has appeared.}
 \label{fig:legled_uk}
 \end{figure}

\begin{figure}[ht]
 \centering
 \includegraphics[scale=.5]{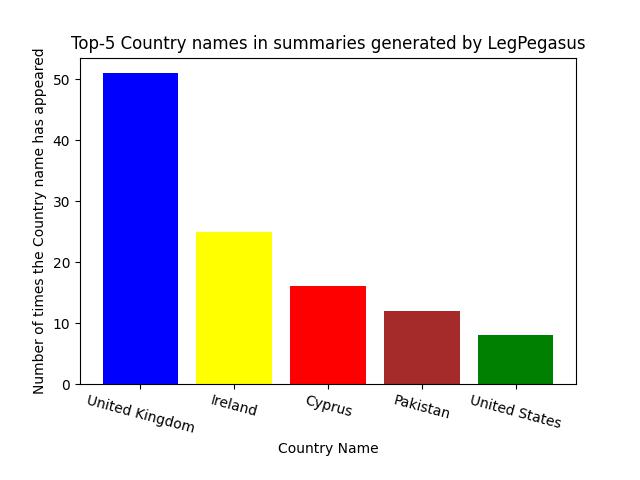}
 \caption{Top-5 country names in the summaries generated by LegPegasus model. x-axis represents the country names in UK-Abs dataset. y-axis represents the number of times a country name has appeared.}
 \label{fig:legpegasus}
 \end{figure}

\begin{figure}[ht]
 \centering
 \includegraphics[scale=.5]{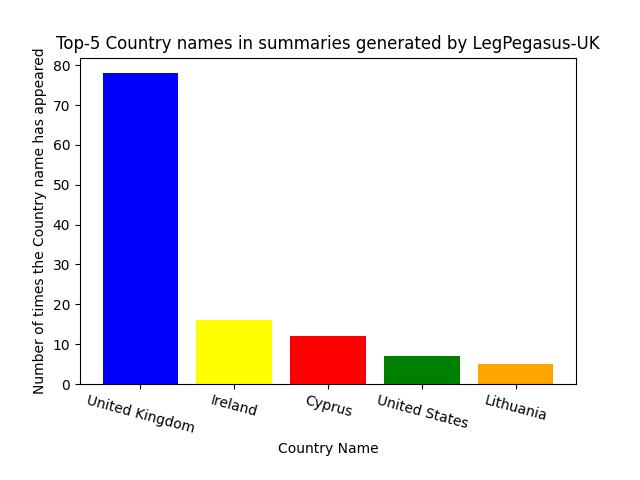}
 \caption{Top-5 country names in the summaries generated by LegPegasus-UK model. x-axis represents the country names in UK-Abs dataset. y-axis represents the number of times a country name has appeared.}
 \label{fig:legpegasus_uk}
 \end{figure}

\begin{figure}[ht]
 \centering
 \includegraphics[scale=.5]{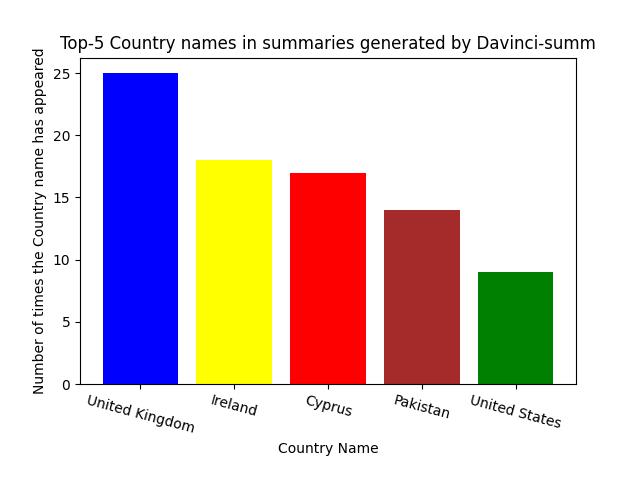}
 \caption{Top-5 country names in the summaries generated by Davinci-summ model. x-axis represents the country names in UK-Abs dataset. y-axis represents the number of times a country name has appeared.}
 \label{fig:davinci_summ}
 \end{figure}

\begin{figure}[ht]
 \centering
 \includegraphics[scale=.5]{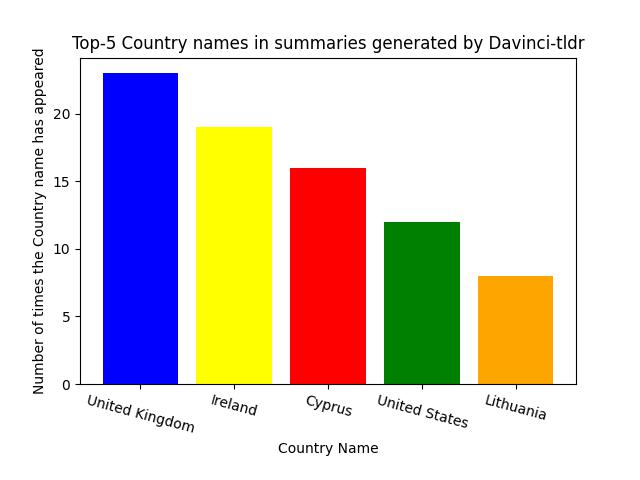}
 \caption{Top-5 country names in the summaries generated by Davinci-TL;DR model. x-axis represents the country names in UK-Abs dataset. y-axis represents the number of times a country name has appeared.}
 \label{fig:davinci-TL;DR}
 \end{figure}

\begin{figure}[ht]
 \centering
 \includegraphics[scale=.5]{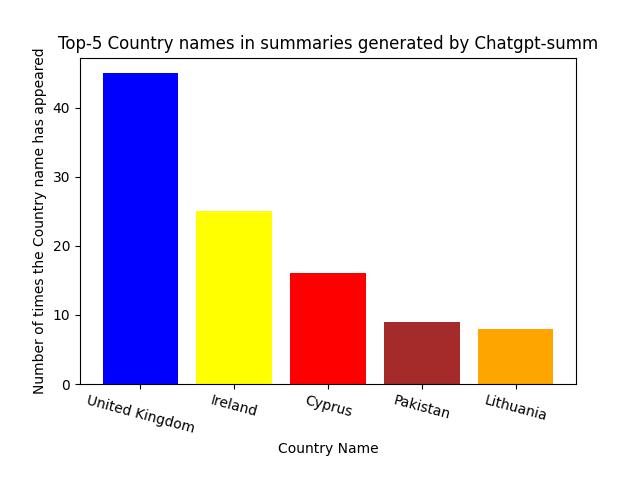}
 \caption{Top-5 country names in the summaries generated by Chatgpt-summ model. x-axis represents the country names in UK-Abs dataset. y-axis represents the number of times a country name has appeared.}
 \label{fig:chatgpt-summ}
 \end{figure}

\begin{figure}[ht]
 \centering
 \includegraphics[scale=.5]{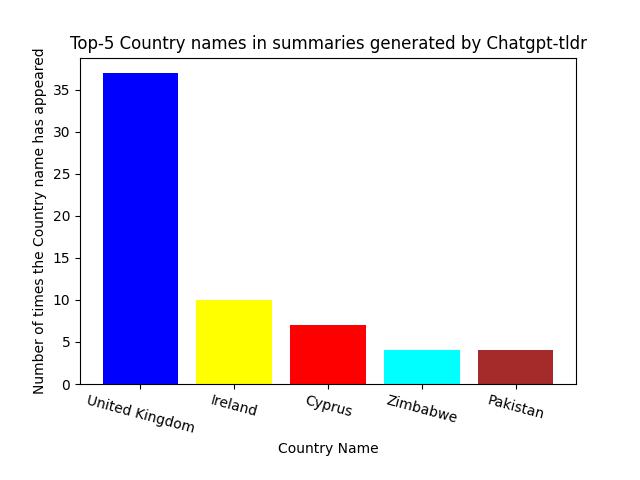}
 \caption{Top-5 country names in the summaries generated by Chatgpt-TL;DR model. x-axis represents the country names in UK-Abs dataset. y-axis represents the number of times a country name has appeared.}
 \label{fig:chatgpt-TL;DR}
 \end{figure}

\begin{figure}[ht]
 \centering
 \includegraphics[scale=.5]{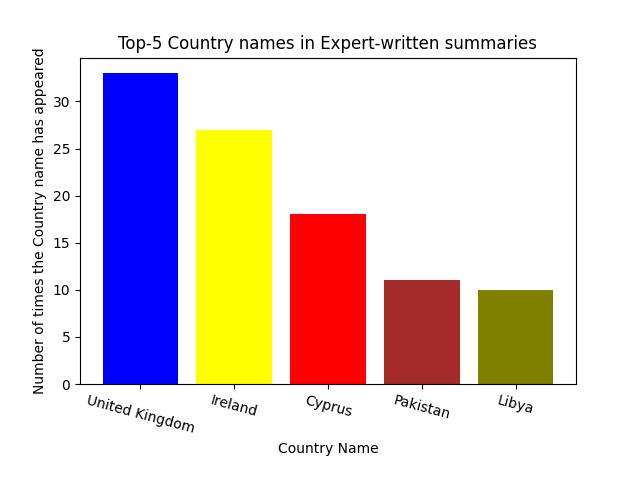}
 \caption{Top-5 country names in the Expert-written summaries. x-axis represents the country names in UK-Abs dataset. y-axis represents the number of times a country name has appeared.}
 \label{fig:Expert}
 \end{figure}

\begin{figure}[ht]
 \centering
 \includegraphics[scale=.5]{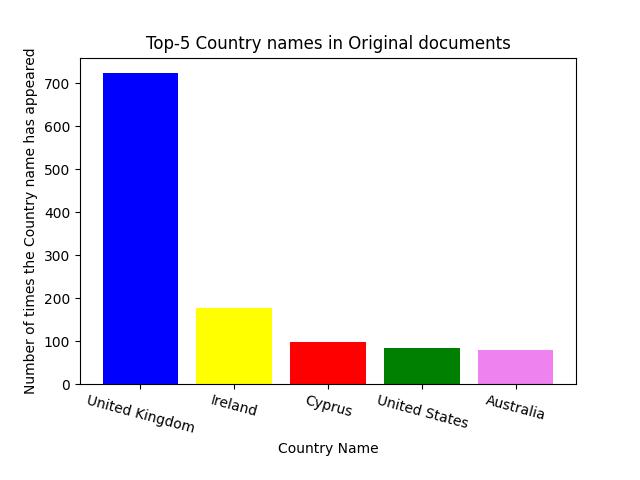}
 \caption{Top-5 country names in the Original documents. x-axis represents the country names in UK-Abs dataset. y-axis represents the number of times a country name has appeared.}
 \label{fig:Doc}
 \end{figure}

Figure ~\ref{fig:legled}, ~\ref{fig:legled_uk}, ~\ref{fig:legpegasus}, ~\ref{fig:legpegasus_uk},
~\ref{fig:davinci_summ}, ~\ref{fig:davinci-TL;DR},
~\ref{fig:chatgpt-summ}, ~\ref{fig:chatgpt-TL;DR},  show the Top-5 country names in the summaries generated by the LegLED, LegLED-UK, LegPegasus, LegPegasus-UK, Davinci-summ, Davinci-TL;DR, Chatgpt-summ and Chatgpt-TL;DR respectively in UK-Abs dataset. Figure ~\ref{fig:Expert} and Figure~\ref{fig:Doc} show the Top-5 countries in the expert-written summaries and original documents in UK-Abs dataset.

We observe that some legal domain-specific abstractive summarization models like LegLED tend to generate some country names like "United States" more than other models specifically due to initial training on US legal data. On the other hand, the LegLED-UK model which was further fine-tuned on UK legal data shows a reduction in the number of times the country name-"United States" has appeared. LegPegasus and LegPegasus-UK models have a lesser number of times the country name-"United States" appearing in their summaries.
Interestingly the general domain LLMs like Chatgpt-TL;DR, Chatgpt-summ, and the expert-written summaries do not have "United States" in the list of Top-5 country names in their generated summaries. Though the original documents have the country name-"United States" in the list of Top-5 country names. Also, Davinci-TL;DR and Davinci-summ have "United States" in the list of Top-5 country names in their generated summaries.

Across all the model-generated summaries, expert-written summaries, and original documents, the most occurring country name is-"United Kingdom" because of the fact that we are working on UK court cases.

Interestingly Country names like Pakistan, Lithuania, and Zimbabwe come up in the list of Top-5 country names in the output summaries generated by the general purpose LLMs as well as the legal domain-specific LLMs.

\begin{figure}[ht]
 \centering
 \includegraphics[scale=.5]{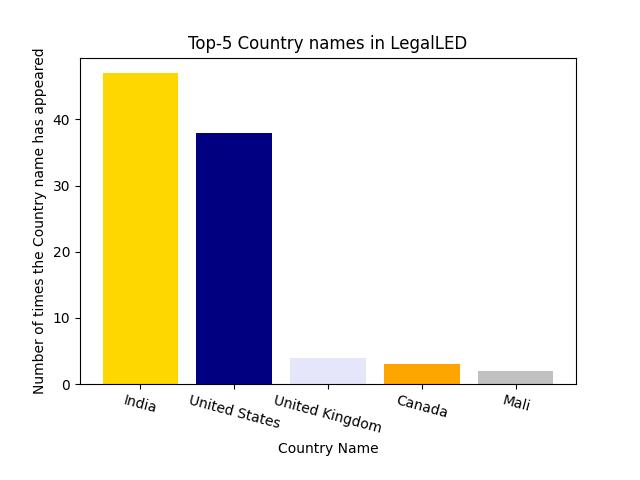}
 \caption{Top-5 country names in the summaries generated by LegLED model in IN-Abs dataset. x-axis represents the country names. y-axis represents the number of times a country name has appeared.}
 \label{fig:legled_in}
 \end{figure}

\begin{figure}[ht]
 \centering
 \includegraphics[scale=.5]{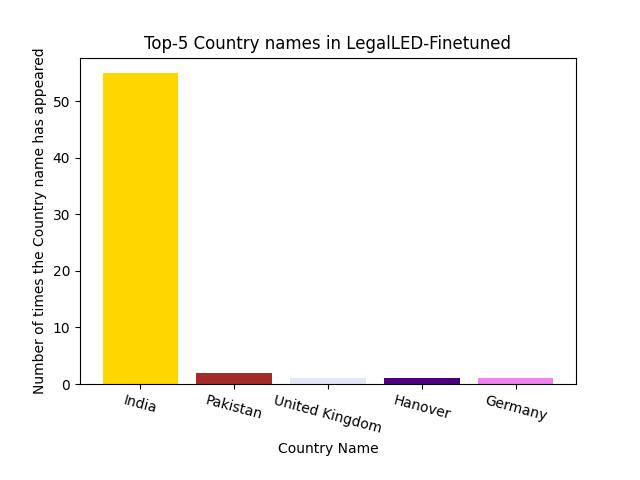}
 \caption{Top-5 country names in the summaries generated by LegLED-IN model in IN-Abs dataset. x-axis represents the country names. y-axis represents the number of times a country name has appeared.}
 \label{fig:legledf_in}
 \end{figure}

\begin{figure}[ht]
 \centering
 \includegraphics[scale=.5]{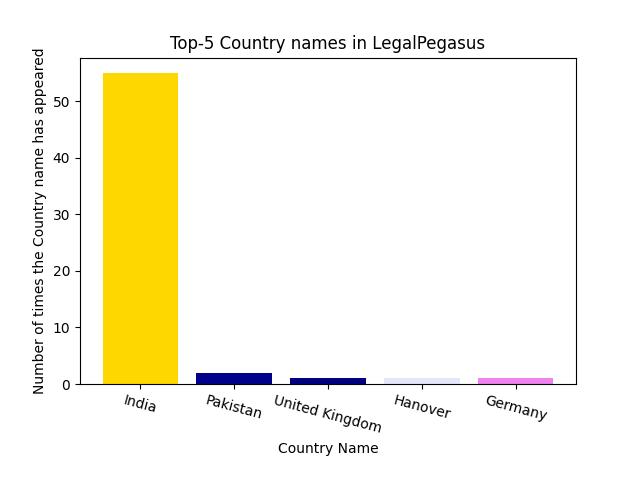}
 \caption{Top-5 country names in the summaries generated by LegPegasus model in IN-Abs dataset. x-axis represents the country names. y-axis represents the number of times a country name has appeared.}
 \label{fig:legpegasus_in}
 \end{figure}

\begin{figure}[ht]
 \centering
 \includegraphics[scale=.5]{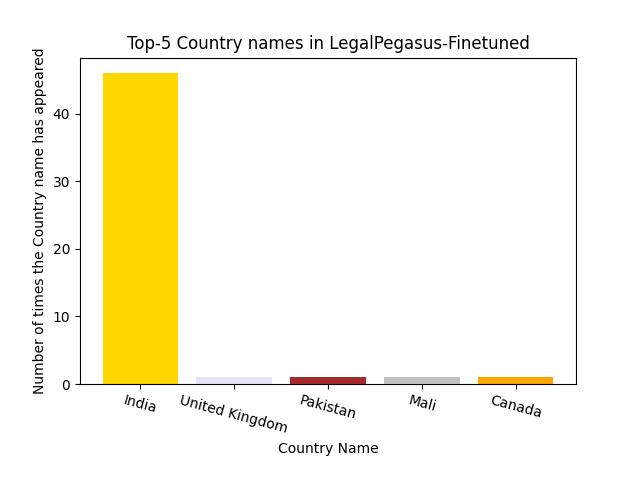}
 \caption{Top-5 country names in the summaries generated by LegPegasus-IN model in IN-Abs dataset. x-axis represents the country names. y-axis represents the number of times a country name has appeared.}
 \label{fig:legpegasusf_in}
 \end{figure}

\begin{figure}[ht]
 \centering
 \includegraphics[scale=.5]{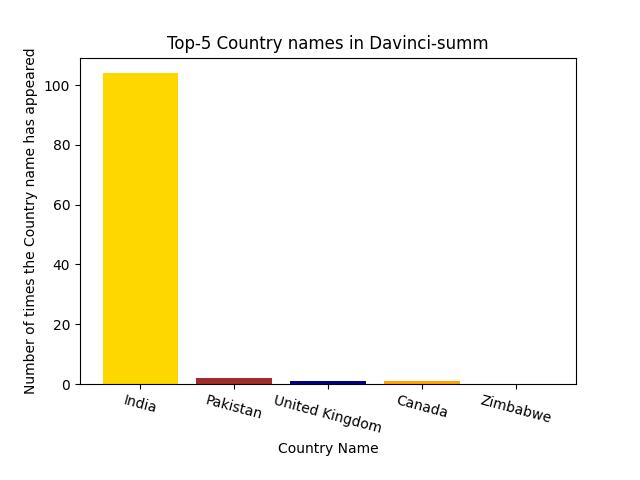}
 \caption{Top-5 country names in the summaries generated by Davinci-summ model in IN-Abs dataset. x-axis represents the country names. y-axis represents the number of times a country name has appeared.}
 \label{fig:davinci_summ_in}
 \end{figure}

\begin{figure}[ht]
 \centering
 \includegraphics[scale=.5]{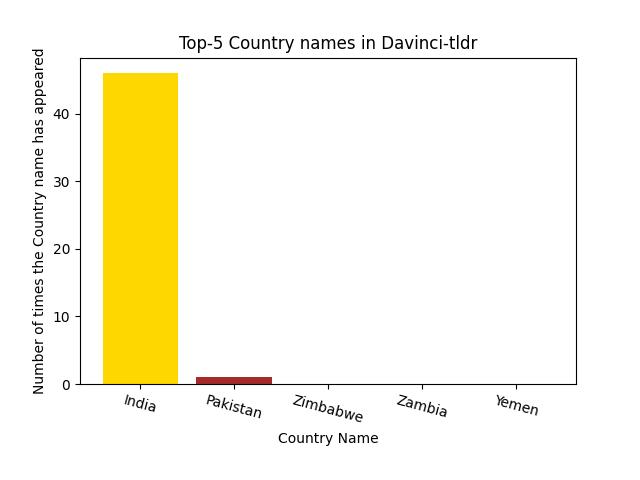}
 \caption{Top-5 country names in the summaries generated by Davinci-TL;DR model in IN-Abs dataset. x-axis represents the country names. y-axis represents the number of times a country name has appeared.}
 \label{fig:davinci-TL;DR_in}
 \end{figure}

\begin{figure}[ht]
 \centering
 \includegraphics[scale=.5]{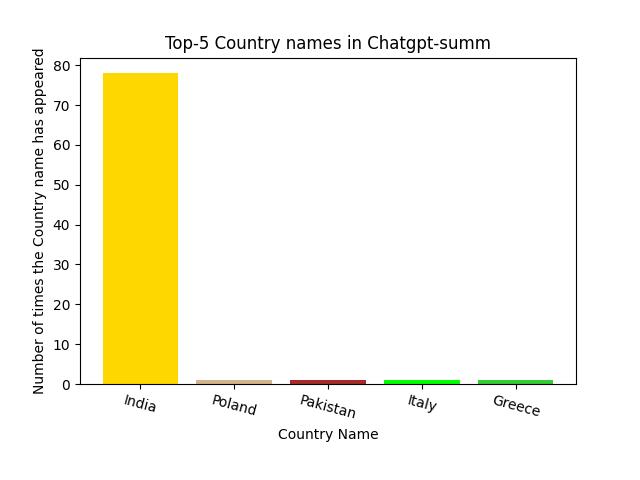}
 \caption{Top-5 country names in the summaries generated by Chatgpt-summ model in IN-Abs dataset. x-axis represents the country names. y-axis represents the number of times a country name has appeared.}
 \label{fig:chatgpt-summ_in}
 \end{figure}

\begin{figure}[ht]
 \centering
 \includegraphics[scale=.5]{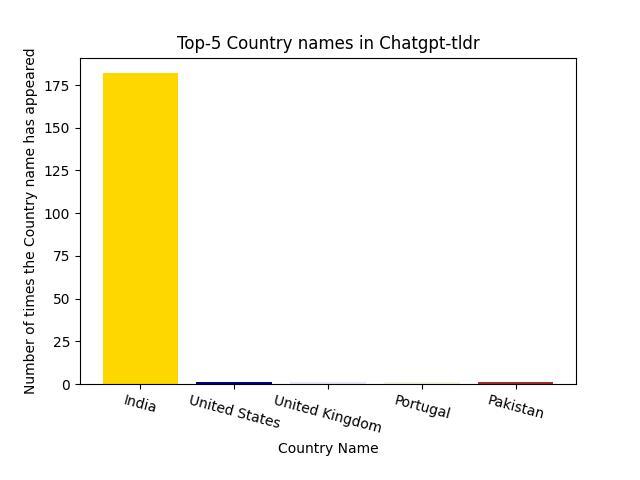}
 \caption{Top-5 country names in the summaries generated by Chatgpt-TL;DR model in IN-Abs dataset. x-axis represents the country names. y-axis represents the number of times a country name has appeared.}
 \label{fig:chatgpt-TL;DR_in}
 \end{figure}

Figure ~\ref{fig:legled_in}, ~\ref{fig:legledf_in}, ~\ref{fig:legpegasus_in}, ~\ref{fig:legpegasusf_in},
~\ref{fig:davinci_summ_in}, ~\ref{fig:davinci-TL;DR_in},
~\ref{fig:chatgpt-summ_in}, ~\ref{fig:chatgpt-TL;DR_in},  show the Top-5 country names in the summaries generated by the LegLED, LegLED-IN, LegPegasus, LegPegasus-IN, Davinci-summ, Davinci-TL;DR, Chatgpt-summ and Chatgpt-TL;DR respectively in IN-Abs dataset. Figure ~\ref{fig:Expert} and Figure~\ref{fig:Doc} show the Top-5 countries in the expert-written summaries and original documents of IN-Abs dataset.
We observe that some legal domain-specific abstractive summarization models like LegLED tend to generate some country names like "United States" more than other models specifically due to initial training on US legal data. On the other hand, the LegLED-UK model which was further fine-tuned on UK legal data shows a reduction in the number of times the country name-"United States" has appeared. LegPegasus and LegPegasus-UK models have a lesser number of times the country name-"United States" appearing in their summaries.
Across all the model-generated summaries, expert-written summaries, and original documents, the most occurring country name is-"India" because of the fact that we are working on Indian court cases. 
The country name-"United Kingdom" is also present in the list of Top-5 Country names in the model-generated summaries of LegLED, LegLED-IN, LegPegasus, LegPegasus-IN, and Chatgpt-TL;DR. Also, Country names like Germany, Italy, Greece, and Pakistan occur in the list of Top-5 country names of several model-generated summaries.

  \begin{figure}[ht]
 \centering
 \includegraphics[scale=.5]{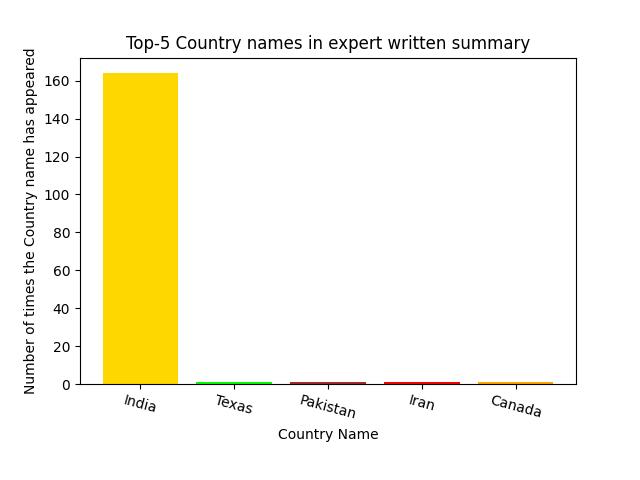}
 \caption{Top-5 country names in the Expert-written summaries in IN-Abs dataset. x-axis represents the country names. y-axis represents the number of times a country name has appeared.}
 \label{fig:Expert_in}
 \end{figure}

  \begin{figure}[ht]
 \centering
 \includegraphics[scale=.5]{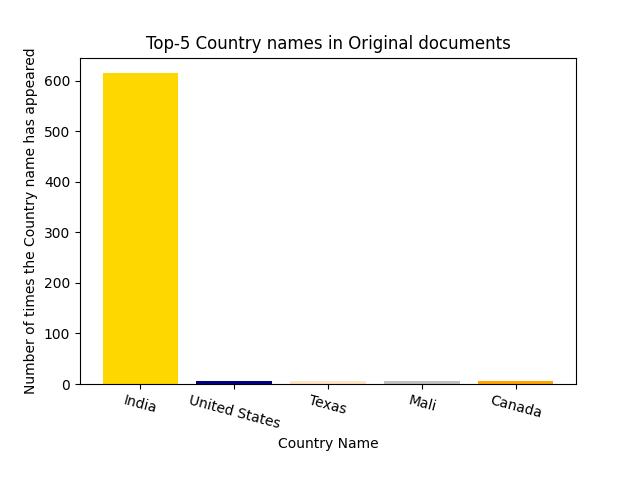}
 \caption{Top-5 country names in the Original documents in IN-Abs dataset. x-axis represents the country names. y-axis represents the number of times a country name has appeared.}
 \label{fig:Doc_in}
 \end{figure}

\textcolor{blue}{ Also, observations show strong biases of certain legal domain-specific abstractive summarization models towards specific country names. Country names like "United States" is generated more by legal domain-specific abstractive models like LegalLED for both IN-Abs and UK-Abs, model-generated summaries. Interestingly Country names like Pakistan, Lithuania, and Zimbabwe come up in the list of Top-5 country names in the output summaries generated by the general purpose LLMs as well as the legal domain-specific LLMs for the UK-Abs dataset. Also, Country names like Germany, Italy, Greece, and Pakistan occur in the list of Top-5 country names of several model-generated summaries for IN-Abs dataset.}

\section{Religious Bias in the outputs of abstractive summarization models}
We study biases wrt to religious-related keywords on the model generated summaries of IN-Abs and UK-Abs dtaset.
We searched for religious keywords like Hindu, Muslim, Christian, Jain, etc in the Original documents, expert-written summaries, and model-generated summaries but we were unable to find such keywords in both IN-Abs and UK-Abs dataset.

We also searched for the act named-Employment Equality (Religion or Belief) Regulations 2003(\url{https://en.wikipedia.org/wiki/Employment_Equality_(Religion_or_Belief)_Regulations_2003}) in the main documents and summaries but the act was not found. 

\textcolor{blue}{We observe no evidences for biases wrt religious keywords in both IN-Abs and UK-Abs model-generated summaries.}

\section{Examples of extracts from the summaries showing crime against women, Race-related keywords, Country names and gender related-keywords}

Table ~\ref{tab:example} shows examples of extracts from the summaries showing crime against women, race-related keywords, country names, and gender related-keywords.

\begin{table*}[tb]
\centering
\begin{tabular}{|p{0.01\textwidth}|p{0.10\textwidth}|p{0.35\textwidth}|p{0.35\textwidth}|}
\hline
\textbf{id} & \textbf{Model} & \textbf{Extract from summary } & \textbf{Explanation}  \\ \hline
\hline
1
&
Chatgpt-TL;DR
&
This is a summary of an argument about the \textcolor{red}{forced marriage} unit in the UK. In response to the revelation of the problem, the Home Office and the Foreign and Commonwealth Office created the \textcolor{red}{Forced Marriage} Unit (FMU) in 2005. The Secretary of State published guidance for those exercising public functions potentially relevant to instances of \textcolor{red}{forced marriage} in 2008. 

&

This is an extract from a summary related to a issue about forced marriage which is a crime against women. 
\\ \hline
2
&
Chatgpt-TL;DR
&
 This case discusses whether Barclays Bank is vicariously liable for the \textcolor{red}{sexual assault}s allegedly committed by the late Dr Gordon Bates on some 126 claimants in this group action.

& 
This is an extract from a summary related to an issue about sexual assault which is a crime against women. 
\\ \hline

3
&
davinci-summ
&
The UK's benefit cap, which restricts the amount of welfare that households can receive, has been ruled discriminatory against single parents and victims of \textcolor{red}{domestic violence}, who are predominantly women.

& 
This is an extract from a summary related to an issue about domestic violence which is a crime against women. 
\\ \hline

4
&
LegLED
&
The Office of the Chief Constable of the Royal Ulster Constabulary today announced that it has concluded that, in treating a \textcolor{red}{black} or \textcolor{red}{female} employee less favourably on racial grounds, the employer acted as he did.

&
This is an extract from a summary related to a issue about black and female employees in their workplace.
\\ \hline

5
&
LegLED
&
The \textcolor{red}{United States} Attorney's Office for the Southern District of New York today announced criminal charges against a \textcolor{red}{South Korean} citizen for his role in a scheme to manipulate the value of the bonds.   

&
This is an extract from the summary which talks about two countries United States and South Korea.
\\ \hline

6
&
davinci-summ
&
The document discusses the issue of whether \textcolor{red}{men} and \textcolor{red}{women} are in the same employment for the purposes of the Equal Pay Act 1970 (now replaced by the Equality Act 2010).

&

This is an extract from a summary talking about male and female equality regarding matters of employment.
\\ \hline

\end{tabular}

\caption{\label{tab:example}
Examples of extracts from the model-generated summaries showing keywords related to crime against women, Race-related keywords, Country names, and gender-related keywords. The keywords are marked in red.}.

\end{table*}

\section{Concluding Discussion}
(I) In our study, we observe slight biases for female-related keywords on both Indian judgement summaries(IN-Abs) and United Kingdom judgement summaries(UK-Abs) for the different General purpose LLMs and pre-trained legal domain-specific abstractive models.

(ii) Also observations show strong biases of certain legal domain-specific abstractive summarization models towards specific country names. Country names like "United States" is generated more by legal domain-specific abstractive models like LegalLED.  Interestingly Country names like Pakistan, Lithuania, and Zimbabwe come up in the list of Top-5 country names in the output summaries generated by the general purpose LLMs as well as the legal domain-specific LLMs for the UK-Abs dataset. Also, Country names like Germany, Italy, Greece, and Pakistan occur in the list
of Top-5 country names of several model-generated summaries for IN-Abs dataset

(iii)The LLM generated summaries for the United kingdom(UK-Abs) dataset show biases towards specific terms related to crime against women. No such strong observation is found for the Indian(IN-Abs) dataset. 

(iv)We do not find strong evidences for biases wrt religious keywords and race-related keywords in our study for both IN-Abs and UK-Abs dataset.

The reasoning behind these biases needs further studies.



\clearpage

\section{Ethics Statement}
\label{sec:ethics}
We are working with general domain LLMs like Text-Davinci-003 and GPT-3.5 Turbo.
For using the OpenAI services for Text-Davinci-003 and GPT-3.5 Turbo, we paid the appropriate prices for using the OpenAI API(\url{https://openai.com/pricing}). 
Models like Legal-Pegasus and Legal-LED
 were used from the Huggingface website at the following links:\url{https://huggingface.co/nsi319/legal-pegasus} and \url{https://huggingface.co/nsi319/legal-led-base-16384}.
 Also, unsupervised extractive summarization models such as Case Summarizer were used from -\url{https://github.com/Law-AI/summarization/tree/aacl/extractive/CaseSummarizer}. Bertsum, a supervised extractive summarization model was used from-\url{https://github.com/nlpyang/BertSum}. SummaRunner which is a supervised and extractive summarization model was used from-\url{https://github.com/hpzhao/SummaRuNNer}.
 
 We have tried to maintain all ethical concerns while performing all the experiments, and we have honestly reported our results in this paper.
 One of the primary ethical concerns in summarizing legal judgments is the accuracy and fairness of the generated summaries. Legal judgments contain complex legal reasoning and nuanced interpretations, and the LLMs tend to simplify, omit, or confuse various pieces of information. It is important to ensure that the generated summaries accurately reflect the key arguments, reasoning, and results of the original judgment. In addition, we should try to reduce bias and inaccuracies in the summaries generated by the LLMs.
 \textcolor{red}{This research focuses on a highly sensitive and distressing issue: crimes against women. We recognize the gravity of this topic and approach it with utmost respect, empathy, and ethical responsibility. Our primary objective is to shed light on the various forms of violence and discrimination faced by women, aiming to raise awareness and promote social change. Please consider very strong words are used as a part of this research, not to hurt the sentiments and feelings of anyone reading this paper. Please be conscious enough while reading this article.}
\clearpage

\bibliographystyle{acl_natbib}
\bibliography{custom,anthology}

\clearpage


\section{Implementation details}
\label{implementation}
\begin{table*}[tb]
\centering
\begin{tabular}{ccc}
\hline
\textbf{Document} & \textbf{Training Time(in hours)} & \textbf{Testing time(in hours)} \\ \hline
\multicolumn{3}{|c|}{\textbf{General-domain LLMs}} \\ \hline

Chatgpt-TL;DR &  & 24  \\ \hline

Chatgpt-summ & & 24 \\ \hline

Davinci-TL;DR &  & 48  \\ \hline

Davinci-summ & & 48 \\ \hline

\multicolumn{3}{|c|}{\textbf{Legal domain-specific abstractive models}} 
\\ \hline

LegLED &  & 1   \\ \hline

LegLED-UK & 24  & 1   \\ \hline

LegLED-IN & 24  & 1   \\ \hline

LegPegasus  & & 1  \\ \hline

LegPegasus-UK  & 24 & 1 \\ \hline

LegPegasus-IN  & 24 & 1 \\ \hline





\end{tabular}

\caption{Training and testing tines of every family of summarization models (approximately in hours)}
\label{tab:train_test}
\end{table*}
Table~\ref{tab:train_test} shows the training and testing times of every family of summarization models. The models were run on one NVIDIA RTX A5000 GPU.

\begin{table*}[tb]
\centering

\begin{tabular}{|p{0.15\textwidth}|p{0.75\textwidth}|}
\hline
\textbf{Model} & \textbf{Hyperparameters} \\ \hline
Chatgpt-TL;DR & max tokens = gold-standard summary length * 1024/Document length , temperature=0.7. \\ \hline

Chatgpt-summ &  max tokens = gold-standard summary length * 1024/Document length. , temperature=0.7  \\ \hline

Davinci-TL;DR &  max tokens = gold-standard summary length * 1024/Document length.\newline  Presence penalty=1.0,  frequency penalty=0.0, temperature=0.7   \\ \hline

Davinci-summ & max tokens = gold-standard summary length * 1024/Document length.\newline  Presence penalty=1.0,  frequency penalty = 0.0, temperature=0.7 
   \\ \hline

LegPegasus & max tokens=gold-standard summary length * 1024/Document length.    \\ \hline

LegPegasus-UK & max tokens=gold-standard summary length * 1024/Document length.    \\ \hline

LegPegasus-IN & max tokens=gold-standard summary length * 1024/Document length.    \\ \hline

LegLED & max tokens=gold-standard summary length * 1024/Document length.  \\ \hline

LegLED-UK & max tokens=gold-standard summary length * 1024/Document length. \\ \hline

LegLED-IN & max tokens=gold-standard summary length * 1024/Document length. \\ \hline
\end{tabular}

\caption{Hyperparameters of the legal domain-specific abstractive models and LLMs used in the work. 'max tokens' indicates the maximum number of words in the summary to be generated for an input chunk of the text of length 1,024 words. Here 'gold-standard summary length' is the actual length (in number of words) of the gold standard summary for the given document.}
\label{tab:hyperparameters}
\end{table*}
Table~\ref{tab:hyperparameters} shows the hyperparameters of the legal domain-specific abstractive models and LLMs used in this work. 

The specific hyperparameter configurations employed to fine-tune the various legal domain-specific abstractive models can be found in Table ~\ref{tab:t5}.


\begin{table*}[tb]
\centering

\begin{tabular}{|p{0.15\textwidth}|p{0.75\textwidth}|}
\hline
\textbf{Model} & \textbf{Finetuning hyperparameters} \\ \hline
       LegPegasus-UK & learning rate: 5e-5, epochs: 2, batch size:1 \\ \hline
         
         LegLED-UK & learning rate: 1e-3, epochs: 3, batch size: 4 \\ \hline

         LegPegasus-IN & learning rate: 5e-5, epochs: 2, batch size:1 \\ \hline
         
         LegLED-IN & learning rate: 1e-3, epochs: 3, batch size: 4 \\ \hline

   
\end{tabular}

\caption{Hyperparameters utilized in finetuning legal domain-specific abstractive models}
\label{tab:t5}
\end{table*}


\end{document}